\documentclass{article}

\usepackage[final]{neurips_2024}

\usepackage{float}
\usepackage{listings}
\usepackage{xcolor}
\colorlet{punct}{red!60!black}
\definecolor{background}{HTML}{EEEEEE}
\definecolor{delim}{RGB}{20,105,176}
\colorlet{numb}{magenta!60!black}
\lstdefinelanguage{json}{
    basicstyle=\normalfont\ttfamily,
    numbers=left,
    numberstyle=\scriptsize,
    stepnumber=1,
    numbersep=8pt,
    showstringspaces=false,
    breaklines=true,
    frame=lines,
    backgroundcolor=\color{background},
    literate=
     *{0}{{{\color{numb}0}}}{1}
      {1}{{{\color{numb}1}}}{1}
      {2}{{{\color{numb}2}}}{1}
      {3}{{{\color{numb}3}}}{1}
      {4}{{{\color{numb}4}}}{1}
      {5}{{{\color{numb}5}}}{1}
      {6}{{{\color{numb}6}}}{1}
      {7}{{{\color{numb}7}}}{1}
      {8}{{{\color{numb}8}}}{1}
      {9}{{{\color{numb}9}}}{1}
      {:}{{{\color{punct}{:}}}}{1}
      {,}{{{\color{punct}{,}}}}{1}
      {\{}{{{\color{delim}{\{}}}}{1}
      {\}}{{{\color{delim}{\}}}}}{1}
      {[}{{{\color{delim}{[}}}}{1}
      {]}{{{\color{delim}{]}}}}{1},
}

\usepackage{graphicx}
\graphicspath{ {./Figures/} }

\usepackage{booktabs}
\usepackage[utf8]{inputenc} % allow utf-8 input
\usepackage[T1]{fontenc}    % use 8-bit T1 fonts
\usepackage{url}            % simple URL typesetting
\usepackage{booktabs}       % professional-quality tables
\usepackage{amsfonts}       % blackboard math symbols
\usepackage{nicefrac}       % compact symbols for 1/2, etc.
\usepackage{microtype}      % microtypography
\usepackage{xcolor}         % colors
\usepackage{amsmath}

\title{THaMES: An End-to-End Tool for Hallucination Mitigation and Evaluation in Large Language Models}

\author{
  Mengfei Liang$^{2}$\thanks{Equal Contributions.} , Archish Arun$^{1,3}$\footnotemark[1] , Zekun Wu$^{1,2}$\thanks{Corresponding Author: p.treleaven@ucl.ac.uk, zekun.wu@holisticai.com} , Cristian Munoz$^1$\\ \textbf{Jonathan Lutch}$^{1,3}$, \textbf{Emre Kazim}$^1$, \textbf{Adriano Koshiyama}$^1$, \textbf{Philip Treleaven}$^{2}$\footnotemark[2]\\\\
  $^1$Holistic AI, $^2$University College London, $^3$Stanford University
}

\begin{document}

\maketitle

\begin{abstract}

Hallucination, the generation of factually incorrect and confabulated content, is a rising issue in the realm of Large Language Models (LLMs). While hallucination detection and mitigation methods exist, they are largely isolated and often inadequate for domain-specific use cases. There is no standardized pipeline combining the necessary components of domain-pertinent dataset generation, hallucination detection benchmarking, and mitigation strategies into one tool. This paper proposes the \textbf{THaMES} framework and library  \footnote{The library is available at \url{https://github.com/holistic-ai/THaMES}.}---a \textbf{T}ool for \textbf{Ha}llucination \textbf{M}itigations and \textbf{E}valuation\textbf{S}. THaMES is an end-to-end solution that evaluates and mitigates hallucinations in LLMs through automated testset generation, multifaceted benchmarking techniques, and flexible mitigation strategies. The THaMES framework is capable of automating testset generation from any corpus of information while achieving high data quality and diversity, maintaining cost-effectiveness through batch processing, weighted sampling, counterfactual validation, and the usage of complex question types. THaMES can also evaluate a model’s capability to identify hallucinations and generate less hallucinated outputs across multiple types of evaluation tasks, including text generation and binary classification. The framework also applies optimal hallucination mitigation strategies tailored to different models and knowledge bases. THaMES contains a variety of hallucination mitigation strategies, including In-Context Learning (ICL), Retrieval Augmented Generation (RAG), and Parameter-Efficient Fine-tuning (PEFT). Evaluating a variety of state-of-the-art LLMs using a knowledge base consisting of academic papers, political news articles, and Wikipedia articles, we find that commercial models such as \texttt{GPT-4o} benefit more from RAG strategies than ICL, and that while open weight models such as \texttt{Llama-3.1-8B-Instruct} and \texttt{Mistral-Nemo} also show improvements with RAG mitigations, they benefit more from the reasoning provided by ICL. In an experiment with open weight model \texttt{Llama-3.1-8B-Instruct}, the PEFT mitigation significantly improved over the base model in aspects of both evaluation tasks.

% \textbf{Keywords:} LLM Hallucination, Question Answering Tasks, Information Retrieval, Privacy and Copyright Enforcement, LLM Bench-marking Tools [.]
\end{abstract}

\section{Introduction and Related Work}

\label{system_diagram}
\begin{figure}
    \centering
\includegraphics[width=1\linewidth]{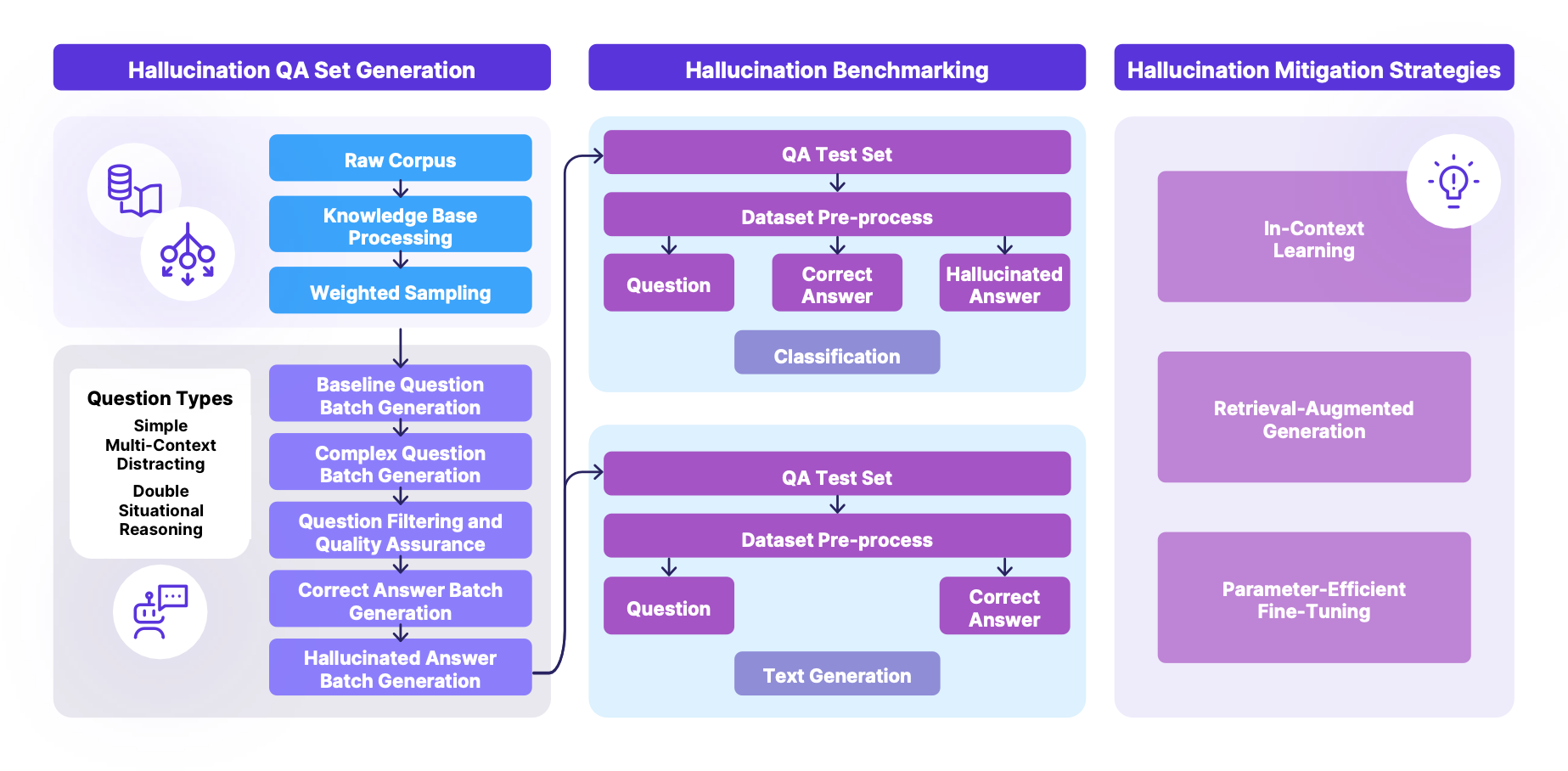}
    \caption{System Diagram of the THaMES Framework, including QA set generation, hallucination benchmarking, and mitigation strategies.}
    \label{fig:system diagram}
\end{figure}

In the development of large language models (LLMs), one persistent issue is hallucination---the generation of outputs which sound plausible but are factually incorrect or unverifiable \citep{Ji_2023}. Research categorizes hallucinations into two types: factuality and faithfulness hallucinations \citep{huang2023factualinconsistencyproblemabstractive}, which arise from sources such as training data, the model’s internal training mechanisms, or inference-time biases. THaMES focuses on question-answering (QA) tasks to capture and evaluate these hallucination types. Although progress has been made with hallucination benchmarks \citep{li2023haluevallargescalehallucinationevaluation, lin-etal-2022-truthfulqa} and mitigation strategies \citep{lewis2021retrievalaugmentedgenerationknowledgeintensivenlp, dhuliawala2023chainofverificationreduceshallucinationlarge}, most studies address specific aspects of the issue without offering a holistic solution. Existing testsets such as those provided by HaluEval \citep{li2023haluevallargescalehallucinationevaluation} and DelucionQA \citep{sadat-etal-2023-delucionqa} provide benchmarks for tasks like QA but rely heavily on time-consuming human annotation. These testsets also lack question complexity and variety. Existing frameworks also tend to focus on a single hallucination evaluation criterion, for example identification or generation individually, which often limits their effectiveness. By our definition, a robust model must satisfy two criteria--to be able to both identify hallucinations and generate text without hallucinations. To solve these issues, we designed THaMES, a \textbf{T}ool for \textbf{Ha}llucination \textbf{M}itigations and \textbf{E}valuation\textbf{S}. THaMES is an end-to-end framework that evaluates and mitigates hallucinations, incorporating batch processing, weighted sampling, and diverse question types, evaluating multiple criteria in order to select optimal mitigation strategies.
The testset generation process in THaMES leverages weighted sampling and refined prompts to create more diverse hallucination testsets. We also introduced a batch generation technique that reduces testset creation costs while improving diversity and completeness. Additionally, THaMES preprocesses test sets and uses downstream tasks to evaluate models on both criteria--hallucination identification and generation. To mitigate hallucinations, multiple advanced strategies such as In-Context Learning (ICL) \citep{dong2024surveyincontextlearning}, which includes Chain-of-Thought (CoT) \citep{wei2023chainofthoughtpromptingelicitsreasoning} and Chain-of-Verification (CoVe) prompting \citep{dhuliawala2023chainofverificationreduceshallucinationlarge} 
, help models reason and verify outputs step-by-step. Retrieval-Augmented Generation (RAG) \citep{lewis2021retrievalaugmentedgenerationknowledgeintensivenlp} improves factual accuracy by grounding responses in external knowledge. Parameter-efficient fine-tuning methods such as LoRA \citep{hu2021loralowrankadaptationlarge} also provide task-specific improvements. Since no single mitigation strategy works best across all models, THaMES evaluates three different strategies, allowing users to select the optimal one based on the model and knowledge base.
We applied the THaMES framework to multiple models, including OpenAI’s \texttt{GPT-4o} \citep{openai2024gpt4technicalreport} and \texttt{GPT-4o-mini}, Meta’s \texttt{Llama-3.1-8B-Instruct} \citep{dubey2024llama3herdmodels}, and Mistral’s \texttt{Mistral-Nemo} \citep{MistralAI}. Results showed that In-Context Learning (CoVe) was less effective for \texttt{GPT-4o}, while RAG significantly improved performance. In contrast, \texttt{Llama-3.1} performed better with In-Context Learning. Additionally, we applied PEFT to the \texttt{Llama-3.1} model, and the results indicated that \texttt{Llama-3.1} produced significantly fewer hallucinated outputs. In summary, The THaMES framework offers a comprehensive solution for hallucination evaluation and mitigation, setting a new standard for reliable LLM development and deployment.

\section{Methodology}

 The THaMES framework is divided into three main components as shown in figure \ref{fig:system diagram}: \textbf{(1)} Testset generation from a user-provided corpus. \textbf{(2)} Baseline metric evaluation based on testset. and \textbf{(3)} Mitigation strategy evaluation based on baseline metrics. In integrating these components, we present an end-to-end solution that transforms raw corpora into specialized LLM benchmarks, revealing optimal hallucination mitigation strategies.

\subsection{Hallucination Benchmark Testset Generation}

This section introduces a method for generating synthetic question-answer (QA) testsets to benchmark hallucination in LLMs. Each QA pair consists of a question, the correct answer, and one hallucinated answer. The QA-set generation process includes seven steps: \textbf{(1)} Knowledge Base Processing, \textbf{(2)} Ground-Truth Weighted Sampling, \textbf{(3)} Baseline Question Batch Generation, \textbf{(4)} Complex Question Batch Evolution, \textbf{(5)} Question Filtering and Quality Assurance, \textbf{(6)} Correct Answer Batch Generation, and \textbf{(7)} Hallucinated Answer Batch Generation.

THaMES is designed to be capable of processing a variety of corpora, so we selected a mix of political news articles, academic papers, and Wikipedia articles to generate our experimental QA testset. In practice, the framework is compatible with various file formats including PDF, TXT, and CSV. We utilized the 
\texttt{VectorStoreIndex} module provided by \texttt{LlamaIndex} \citep{Liu_LlamaIndex_2022} to build a knowledge base out of the raw corpus, selecting text blocks (nodes) and semantically similar neighbors to generate questions and answers. Many existing frameworks use basic random sampling to select text nodes as supporting context for question generation. In implementing THaMES, we designed a weighted random sampling method to ensure a balanced and diverse text block sampling. The sampling probability for each text node $i$ is calculated as: $p_{i}=\frac{w_{i}}{\sum_{j=1}^{n} w_{j}}$, where $w_i=\frac{1}{c_i+1}$. This approach, combined with the \texttt{text-embedding-large-3} embedding model, improved corpus coverage and generated a more diverse testset, tested by comparing the number of observed node retrievals with a uniform retrieval distribution (see Appendix \ref{sampling}). 

We chose \texttt{GPT-4o-mini} \citep{openai2024gpt4technicalreport} for question generation to minimize human validation, due to its high performance and relatively low cost. The HaluEval testset had simple questions with short answers, so we created six question types: \texttt{[simple, reasoning, multi-context, situational, distracting, and double]} based on RAGAS \citep{es2023ragasautomatedevaluationretrieval} and Giskard \citep{giskard}. Unlike these frameworks, our type definitions were designed not for retrieval-augmented generation (RAG) systems but for general LLM hallucination evaluation. We also introduced rules to ensure high generated question quality, such as avoiding ambiguous references, numerical estimates, ensuring self-containment, etc. (detailed in Appendix \ref{question_ruleset}). To generate \texttt{simple}-type questions, we used few-shot prompting, instructing \texttt{GPT-4o-mini} to output batches of JSON objects (Appendix \ref{baselinejsonoutputformat}). Batch generation was chosen for its cost-effectiveness and ability to minimize repeated questions, ensuring testset diversity. The questions were then evolved into one of the six predefined types through a filtering process (Appendix \ref{evolution_prompts}) to ensure quality and compliance with our criteria. Next, the model generated correct answers for each question, using the original context from the knowledge base to ensure factual accuracy.

A key feature of the THaMES testset is its inclusion of hallucinated answers. HaluEval \citep{li2023haluevallargescalehallucinationevaluation} used other models including \texttt{GPT-3.5} \citep{brown2020languagemodelsfewshotlearners} to generate and select hallucinated answers, but this lacked interpretability and did not guarantee the selection of the most distracting answer. THaMES improves this by using fine-tuned NLI (\texttt{deberta-v3-base-tasksource-nli}) \citep{sileo2023tasksource} and hallucination evaluation models (\texttt{HHEM-2.1-Open}) \citep{tang2023understandingfactualerrorssummarization, niu2024ragtruthhallucinationcorpusdeveloping, luo2023chatgptfactualinconsistencyevaluator} to assess the Entailment and Factual Consistency Scores of generated answers. These scores are combined to form an Ensemble Score: $\texttt{Ensemble Score} = \texttt{Entailment Score} + \texttt{Factual Consistency Score}$. Lower scores indicate stronger hallucinations, helping to select the most hallucinated answers. Figure \ref{fig:system diagram} provides more details on the generation process with a system flowchart. Finally, we used the selected corpus to generate a total of 2,100 sets of data (300 sets for each type of question). Each set of data includes a question, a correct answer, and the best hallucinated answer.

\subsection{Hallucination Evaluation}
In this section, we conduct a comprehensive analysis of a benchmark designed to evaluate the ability of LLMs to handle hallucinations in QA tasks. We selected a range of mainstream LLMs for testing, including \texttt{GPT-4o (05-13-2024)}  and \texttt{GPT-4o-mini (07-18-2024)} from OpenAI's GPT series \citep{openai2024gpt4technicalreport} [deployed through Azure Cloud Platform], the open-source Llama series \citep{dubey2024llama3herdmodels}, and Mistral-Nemo \citep{MistralAI}. Due to computational resource constraints, we were unable to test the full-parameter versions of these models, so we opted to use smaller, quantized versions of the open-source models with Ollama \citep{ollama2024}. Additionally, for the fine-tuned model, we loaded \texttt{Llama-3.1-8B-Instruct} from HuggingFace. For details on the LoRA fine-tuning configuration, see Appendix \ref{lora_parameters}.

We employed two sets of metrics for evaluation. The first set includes metrics based on those defined by RAGAS \citep{es2023ragasautomatedevaluationretrieval}: answer faithfulness, relevancy, semantic similarity, and correctness. Here, we present the formula used to calculate answer relevancy: 
% \begin{equation}
    $\texttt{answer relevancy} = \frac{1}{N} \sum_{i=1}^{N} \frac{E_{g_i} \cdot E_o}{\|E_{g_i}\|\|E_o\|}$
% \end{equation}
where $E_{g_i}$ is the embedding of the generated question $i$, $E_o$ is the embedding of the original question, $N$ is the number of generated questions. Formulas for other metrics are provided in the Appendix \ref{additional formulas}. We calculated these metrics by comparing the model-generated answers with the ground truth answers in our testset. We derived our second set of metrics from HaluEval \citep{li2023haluevallargescalehallucinationevaluation}, where the LLM is randomly presented with either the correct or hallucinated answer, each with a $50\%$ probability, to assess the model's ability to identify hallucinations. The model's performance is evaluated using accuracy, calculated as follows: $\texttt{Accuracy} = \frac{\texttt{\# Correct Predictions}}{\texttt{\# Total Predictions}}$. This helps determine how well the model distinguishes between correct and hallucinated answers given a question.

\subsection{Hallucination Mitigation}
In this section, we introduce various hallucination mitigation techniques utilized by THaMES: In-Context Learning (ICL) \citep{dong2024surveyincontextlearning} methods such as Chain-of-Verification (CoVe) \citep{dhuliawala2023chainofverificationreduceshallucinationlarge}, Retrieval-Augmented Generation (RAG) \citep{lewis2021retrievalaugmentedgenerationknowledgeintensivenlp}, and parameter-efficient fine-tuning (PEFT) \citep{peft} strategies.

Our first chosen mitigation strategy was In-Context Learning \citep{dong2024surveyincontextlearning}, which generalizes approaches that use prompting techniques to elicit complex reasoning from the model. For experimental purposes, we chose Chain-of-Verification \citep{ritunchainofverification2024} as our ICL method, which adds a verification step to improve factual accuracy. Implementing CoVe involves prompting the LLM at multiple stages. After generating the baseline response, the model is prompted to generate task-specific verification questions. The model is then prompted again to answer these questions independently from the initial reasoning. This independence helps reduce bias and allows the model to evaluate and refine its own output critically.  

THaMES also includes a RAG strategy \citep{lewis2021retrievalaugmentedgenerationknowledgeintensivenlp}, grounding the LLM’s responses in external knowledge. Our chosen method in building the RAG knowledge base was to collect cases where the model incorrectly classified answers in a baseline hallucination identification evaluation. When encountering a never-before-seen question, the model can search its knowledge base for similar questions that it failed to answer correctly in prior evaluations. Injecting these failed cases directly into the prompt input provides few-shot \citep{brown2020languagemodelsfewshotlearners} formatted context to the model, potentially enabling it to avoid generating or incorrectly classifying hallucinations similar to cases it has seen before.

Additionally, we examined parameter-efficient fine-tuning (PEFT) \citep{peft} methods, specifically LoRA \citep{hu2021loralowrankadaptationlarge}, to fine-tune LLMs for better hallucination detection performance. Inspired by in-context knowledge editing (IKE) \citep{zheng2023editfactualknowledgeincontext}, we used the same collection of failed QA pairs from baseline evaluation to create a training set, utilizing a prompt-tuning method to fine-tune the models to learn the correct responses to these questions. LoRA receives an additional set of parameters that contain the newly learned knowledge and can be directly concatenated with the original model parameters. Through such fine-tuning, LLMs can generate more accurate answers and reduce hallucinated output. However, PEFT requires access to the parameters of the base model, as well as large computational costs to run inference. Due to these factors, we were only able to run one experiment using PEFT as a mitigation technique. For our experiment, we chose to use \texttt{Llama-3.1-8B-Instruct} as the base model and ran the text generation evaluation. Despite this, the THaMES framework is adaptable to a wide range of models and fine-tuning, provided with a model's base parameters.

The selection and combination of these techniques form THaMES, a robust end-to-end domain-flexible framework for LLM hallucination analysis, from test set generation to benchmarking, to hallucination mitigation.

\section{Discussion and Results}
We conducted comprehensive experiments under two hallucination evaluation tasks, comparing the effects of three different mitigation strategies on model performance. The results, as shown in table \ref{tab:table1}, indicate that while overall performance improved after applying mitigation strategies, no optimal strategy works across all models. For example, we found that for GPT-4o, the prompt-based In-Context Learning (ICL) strategy had a limited effect on improving the model, suggesting that its reasoning capabilities are already high and have reached a performance bottleneck. In contrast, using RAG, which introduces external knowledge, significantly enhanced the model’s ability to prevent hallucinations. Open-weight models like Llama-3.1 showed different behavior with these strategies. While RAG helped reduce hallucination generation in \texttt{Llama-3.1-8B-Instruct}, ICL notably improved its accuracy in detecting hallucinations.  With these evaluations in mind, the system we designed allows for the application of different mitigation strategies tailored to the specific model and task, enabling the models to achieve optimal performance in detecting or producing less hallucination. Due to computational restraints, we were only able to conduct fine-tuning mitigation (PEFT) evaluations on the \texttt{Llama-3.1-8B-Instruct} model. Despite these limitations, results show significant improvement over the original model in text generation, specifically in regard to Answer Relevancy, Answer Correctness, and Answer Similarity. Additionally, in Hallucination Identification, we saw improvement over the baseline model in Recall and overall F-1 score. This highlights the potential of fine-tuning in reducing hallucinated output.

\begin{table}[H]
  \centering
  \caption{Comparison of models on Hallucination Generation and Identification. The data is split into two hallucination criteria--Hallucination Generation and Hallucination Identification--with each model undergoing several mitigation experiments: Original, ICL, RAG and PEFT (applied only to Llama3.1). The metrics $A_F$, $A_R$, $A_C$, and $A_S$ represent Answer Faithfulness, Answer Relevancy, Answer Correctness, and Answer Similarity, respectively. The metrics $Acc.$, $Prec.$, $Rec.$, and $F1$ represent Accuracy, Precision, Recall, and F1-Score, respectively.}
  \label{tab:table1}
  \begin{tabular}{lcccccccc}
    \toprule
    Model & \multicolumn{4}{c}{Text Generation} & \multicolumn{4}{c}{Hallucination Identification} \\
    \cmidrule(r){2-5} \cmidrule(r){6-9}
     & $A_F$ & $A_R$ & $A_C$ & $A_S$ & $Acc.$ & $Prec.$ & $Rec.$ & $F1$ \\
    \midrule
    GPT-4o & & & & & & & & \\
    \midrule
    Original   & 0.520 & 0.761 & 0.692 & 0.785  & 0.608 & 0.582 & 0.719 & 0.644 \\
    ICL        & 0.525 & 0.759 & 0.691 & 0.785  & 0.600 & 0.585 & \textbf{0.719} & \textbf{0.645} \\
    RAG        & \textbf{0.530} & \textbf{0.764} & \textbf{0.699} & \textbf{0.785} & \textbf{0.624} & \textbf{0.691} & 0.459 & 0.552 \\
    \midrule
    GPT-4o-mini & & & & & & & & \\
    \midrule
    Original   & \textbf{0.439} & 0.785 & \textbf{0.688} & \textbf{0.802} & 0.588 & 0.726 & 0.242 & 0.363 \\
    ICL        & 0.431 & \textbf{0.787} & 0.683 & 0.801 & 0.588 & 0.724 & 0.232 & 0.351 \\
    RAG        & 0.400 & 0.772 & 0.674 & 0.801 & \textbf{0.688} & \textbf{0.730} & \textbf{0.666} & \textbf{0.696} \\
    \midrule
    Mistral-Nemo & & & & & & & & \\
    \midrule
    Original   & 0.306 & \textbf{0.411} & 0.159 & 0.077 & 0.436 & 0.405 & 0.304 & 0.347 \\
    ICL        & \textbf{0.314} & 0.407 & \textbf{0.159} & \textbf{0.077} & 0.455 & 0.392 & \textbf{0.308} & 0.345 \\
    RAG        & 0.307 & 0.407 & 0.158 & 0.076 & \textbf{0.506} & \textbf{0.515} & 0.277 & \textbf{0.361} \\
    \midrule
    Llama-3.1-8B-Instruct & & & & & & & & \\
    \midrule
    Original   & 0.397 & 0.452 & 0.479 & 0.701 & 0.528 & 0.525 & 0.737 & 0.613 \\
    ICL        & \textbf{0.403} & 0.455 & 0.484 & 0.695 & \textbf{0.550} & \textbf{0.554} & 0.752 & 0.638 \\
    RAG        & 0.402 & 0.460 & 0.491 & 0.694 & 0.515 & 0.508 & 0.883 & 0.645 \\
    PEFT  & 0.234 & \textbf{0.752} & \textbf{0.567} & \textbf{0.723} & 0.498 & 0.497& \textbf{1.000} & \textbf{0.664}\\
    \bottomrule
  \end{tabular}
\end{table}

\paragraph{Limitations and Future Work}

While THaMES fills a critical gap in current hallucination research by providing a comprehensive framework, it still has some limitations. \textbf{(1)} First, due to limited computational resources, we were only able to experiment with quantized and small-parameter versions of the models, which constrained the effectiveness of our mitigation methods. \textbf{(2)} Secondly, in the dataset construction process, we relied heavily on GPT-4o-mini, which, despite being cost-optimized, still incurred large costs and the dataset quality is limited by the model. \textbf{(3)} Lastly, due to resource constraints, we were unable to fully explore LoRA fine-tuning techniques to achieve optimal hallucination mitigation. For future work, we aim to further optimize the dataset generation process to make it even more cost-effective. Additionally, incorporating small levels of human validation and feedback steps could improve the quality of our datasets. Finally, we plan to extend the THaMES framework to support more downstream tasks, such as text summarization, broadening its applicability.

% \newpage
\bibliographystyle{plainnat}
\bibliography{reference}

\appendix
\section{Appendix / supplemental material} 
\subsection{Question Generation}
\subsubsection{Question Types}
\label{question_types}
To ensure a high amount of question diversity within the generated QA-pair test set, we selected 6 types of question format, based on concepts defined by the RAGAS \citep{es2023ragasautomatedevaluationretrieval} and Giskard \citep{giskard} frameworks

\begin{itemize}
    \item \textbf{Simple:} Basic questions that do not require complex reasoning or multiple contexts. (Note: Giskard \citep{giskard} explicitly defines these as being generated from a single piece of the knowledge base.)
    \item \textbf{Reasoning:} Questions designed to enhance the need for reasoning to answer them effectively (at least one leap of intuition required to correlate the answer to the correct information from the knowledge base).
    \begin{itemize}
        \item \textbf{Multi-Context:} Questions that necessitate information from multiple related sections or chunks to formulate an answer.
    \end{itemize}
    \item \textbf{Situational:} Questions that include additional context to evaluate the ability of the model to produce relevant answers according to the context.
    \begin{itemize}
        \item \textbf{Distracting:} Questions made to confuse the model's reasoning ability with a distracting element from the knowledge base that is irrelevant to the question. 
    \end{itemize}
    \item \textbf{Double Questions:} Questions with two distinct parts and answers, with or without connection to one another.
\end{itemize}

\subsection{Question Generation Prompts}
\subsubsection{Baseline Output Structure}
We used the following structured output format for few-shot prompting baseline question generation. 
\label{baselinejsonoutputformat}
\begin{lstlisting}[language=json,numbers=none]
[
    {
        "question": "In what year did Jann Wenner complete secondary education at Chadwick School?",
        "answer": "Jann Wenner completed secondary education at Chadwick School in 1963."
    },
    ... // more question-answer pairs follow
]
\end{lstlisting}

\subsubsection{Simple Question Generation Prompt}
\label{simple_generation_prompt}
\begin{lstlisting}[language=json,numbers=none]

Your role is to generate evaluation questions based a knowledge base, a subsection of which is provided to you as a list of context paragraphs. We are building a testset for a model. Assume the model that is being tested cannot see the context when answering the question.

Your question must be related to a provided context.  
Please respect the following rules to generate the question:
- The answer to the question should be found inside the provided context
- The question must be self-contained - in other words, the question must explicitly state any necessary titles, names, or terms. Assume that the model to be tested with this testset has no prior knowledge of the context. Avoid using pronouns or references that are not explicitly mentioned in the context. Additionally, there must be a singular and clear answer to the question. Ensure that the potential answer to the question has not changed since the context was provided.
- The question and answer must be in this language: {language}

The user will provide the context, consisting of multiple paragraphs delimited by dashes "------".
You will return a question based exclusively on the provided context. 
You will output a list of {num_questions} JSON objects with key 'question' mapping to the generated question, without any other wrapping text or markdown. Ensure that you will only return valid JSON for example:
[{{"question": "xx", "question_type": "simple"}}, {{"question": "xx", "question_type": "simple"}}, ...

\end{lstlisting}

\subsubsection{Reasoning Question Evolution Prompt}
\label{evolution_prompts}
\begin{lstlisting}[language=json,numbers=none]
 """Your role is to modify evaluation questions generated from a knowledge base, a subsection of which is provided to you as a list of context paragraphs. You will be given simple questions that need to be modified to include complex reasoning and logical connections between different pieces of data within the provided context. Evolve the provided simple questions into questions into questions designed to enhance the need for reasoning to answer them effectively (at least one leap of intuition required to correlate the answer to the correct information from the knowledge base)

You will be given an input of questions that looks like the following: 
[{{"question": "xx1", "question_type":"simple"}}, {{"question": "xx2", "question_type":"simple"}}, ...]
as well as some context paragraphs.
The user will provide the context, consisting of multiple paragraphs delimited by dashes "------".
You will output a list of JSON objects with the same length as the input, with key 'question' mapping to the modified question, without any other wrapping text or markdown. Additionally, modify the question_type JSON parameter for all of these questions to be 'reasoning'. Ensure that you will only return valid JSON for example:
[{{"question": "modified xx1 to include complex reasoning", "question_type": "reasoning"}}, {{"question": "modified xx2 to include complex reasoning", "question_type": "reasoning"}}, ...].\n""",


\end{lstlisting}

\subsubsection{Multi-Context Question Evolution Prompt}
\begin{lstlisting}[language=json,numbers=none]
xe      
"""Your role is to modify evaluation questions generated from a knowledge base, a subsection of which is provided to you as a list of context paragraphs. You will be given simple questions that need to be modified to have non-deterministic layers, potentially created from additionally provided context. The questions should require the integration of information from multiple sources or sections within the knowledge base, and the modifications should lead the model to use comprehension to get to the correct solution, rather than straight information retrieval. Evolve the provided simple questions into questions that necessitate information from multiple related sections or chunks of knowledge to formulate a single answer.
    
You will be given an input of questions that looks like the following: 
[{{"question": "xx1", "question_type":"simple"}}, {{"question": "xx2", "question_type":"simple"}}, ...]
as well as some context paragraphs. 
The user will provide the context, consisting of multiple paragraphs delimited by dashes "------". The additional context will be provided subsequently, in the same format.
You will output a list of JSON objects with the same length as the input, with key 'question' mapping to the modified question, without any other wrapping text or markdown. Additionally, modify the question_type JSON parameter for all of these questions to be 'multi_context'. Ensure that you will only return valid JSON for example:
[{{"question": "modified xx1 to include multiple contexts", "question_type":"multi_context"}}, {{"question": "modified xx2 to include multiple contexts",  "question_type":"multi_context"}}, ...].\n""",


\end{lstlisting}

\subsubsection{Situational Question Evolution Prompt}
\begin{lstlisting}[language=json,numbers=none]

"""Your role is to modify evaluation questions generated from a knowledge base, a subsection of which is provided to you as a list of context paragraphs. You will be given simple questions that need to be modified to have situational context about the subject matter inside the question. 
Please respect the following rules to generate the question:
- The question must include information from the situational context.
- The question must sound plausible and coming from a real human user.
- The question can start with any form of greeting or not, choose randomly
- The original question and answer should be preserved.
- The question must be self-contained and understandable by humans. 
- The question must be in this language: {language}.        

You will be given an input of questions that looks like the following: 
[{{"question": "xx1", "question_type":"simple"}}, {{"question": "xx2", "question_type":"simple"}}, ...]
as well as some context paragraphs. 
The user will provide the context, consisting of multiple paragraphs delimited by dashes "------". The additional context from which to generate the situational statements will be provided subsequently, in the same format.
You will output a list of JSON objects with the same length as the input, with key 'question' mapping to the modified question, without any other wrapping text or markdown. Additionally, modify the question_type JSON parameter for all of these questions to be 'situational'. Ensure that you will only return valid JSON for example:
[{{"question": "modified xx1 to include additional statements", "question_type": "situational"}}, {{"question": "modified xx2 to include additional statements", "question_type":"situational"}}, ...].\n"""


\end{lstlisting}

\subsubsection{Distracting Question Evolution Prompt}
\begin{lstlisting}[language=json,numbers=none]

"""Your role is to modify evaluation questions generated from a knowledge base, a subsection of which is provided to you as a list of context paragraphs. You will be given simple questions that need to be modified to include statements that describe aspects of the additional context that are unrelated to the question. While the initial question must remain preserved, the additional context should introduce elements meant to confuse the LLM and retrieval from the knowledge base, evaluating the model's ability to focus on the relevant information for the correct answer. Evolve the provided simple questions into questions made to confuse the retrieval part of a model's RAG with a distracting element from the knowledge base but irrelevant to the question. (Designed to mess with embedding engines - leaves more reasoning work for the LLM)

You will be given an input of questions that looks like the following: 
[{{"question": "xx1", "question_type":"simple"}}, {{"question": "xx2", "question_type":"simple"}}, ...]
as well as some context paragraphs. 
The user will provide the context, consisting of multiple paragraphs delimited by dashes "------". The additional context from which to generate the distracting statements will be provided subsequently, in the same format.
You will output a list of JSON objects with the same length as the input, with key 'question' mapping to the modified question, without any other wrapping text or markdown. Additionally, modify the question_type JSON parameter for all of these questions to be 'distracting'. Ensure that you will only return valid JSON for example:
[{{"question": "modified xx1 to include distracting statements", "question_type":"distracting"}}, {{"question": "modified xx2 to include distracting statements", "question_type":"distracting"}}, ...].\n"""


\end{lstlisting}

\subsubsection{Double Question Evolution Prompt}
\begin{lstlisting}[language=json,numbers=none]

"""Your role is to modify evaluation questions generated from a knowledge base, a subsection of which is provided to you as a list of context paragraphs. You will be given simple questions that need to be modified to include two distinct parts, each requiring a different piece of information from the provided context. The questions should be compound queries that consist of two distinct parts, evaluating the capabilities of the query rewriter of the RAG to accurately address both parts of the question using the provided context. You may generate the second part of the question based on the additional context provided. Evolve the provided simple questions into questions with two distinct parts to evaluate the capabilities of the query rewriter of the RAG. These questions should have 2 different answers or 2 different parts of the same answer.

You will be given an input of questions that looks like the following: 
[{{"question": "xx1", "question_type":"simple"}}, {{"question": "xx2", "question_type":"simple"}}, ...]
as well as some context paragraphs. 
The user will provide the context, consisting of multiple paragraphs delimited by dashes "------". The additional context from which to generate the second part of the question will be provided subsequently, in the same format.
You will output a list of JSON objects with the same length as the input, with key 'question' mapping to the modified question, without any other wrapping text or markdown. Additionally, modify the question_type JSON parameter for all of these questions to be 'double'. Ensure that you will only return valid JSON for example:
[{{"question": "modified xx1 to include 2nd part", "question_type": "double"}}, {{"question": "modified xx2 to include 2nd part", "question_type":"double"}}, ...].\n"""


\end{lstlisting}

\subsection{Question Quality Assurance}
\subsubsection{High-Quality Question Criteria}
\label{question_ruleset}
\begin{enumerate}
    \item Questions must not refer to any ambiguous events, contexts, articles, or reports. 
    \item Questions must be self-contained.
    \item Questions cannot contain numerical estimates. 
    \item Each question must stand on its own without previous questions. If a batch contains two questions about the same topic, each question must be self-contained, without relying on the other question for context. 
    \item Questions should be relevant to the provided context.  
\end{enumerate}
%  The complete prompt including the quality-assurance criteria can be found in Appendix \ref{evolution criteria}

\subsubsection{Additional Question Evolution Criteria} \label{evolution criteria}
\begin{lstlisting}[language=json,numbers=none]
PROMPT_EVOLUTION_ADDITIONAL_CRITERIA = """
Your generated questions must meet the following criteria:
-<IMPORTANT> Do not use words that reference the context directly, such as "this", "that", "it", etc. Additionally, do not include phrases such as "as mentioned in the context" or "according to the passage" in the question. Assume the model that is being tested cannot see the context when answering the question.</IMPORTANT>
- The answer to the question should be found inside the provided context
- The question must be self-contained - in other words, the question must explicitly state any necessary titles, names, or terms. Assume that the model to be tested with this testset has no prior knowledge of the context. Avoid using pronouns or references that are not explicitly mentioned in the context. Additionally, there must be a singular and clear answer to the question.
- Ensure that the modifications you make to the questions lead to and preserve a fixed and singular answer to said question, and that the potential answer to the question has not changed since the context was provided.
- When generating questions with numerical answers, ensure that the number is relatively easy to verify, without margin for interpretation. For example, if the answer is a date, ensure that the date is clearly stated in the context. If the answer is a large number, ensure that the number is explicitly mentioned in the context. Assume that the model cannot see the context, and the number in question can be generally understood without additional context.
- The questions should be specific and unambiguous, with a clear and singular answer. Avoid generating questions that are vague, open-ended, or have multiple possible answers. Ensure that the question is clear and concise, with a single correct answer that can be found in the provided context.
- The questions should be relevant to the context provided, with the answer being directly supported by the information in the context. Ensure that the question is closely related to the information in the context, and that the answer can be found within the context paragraphs.
- Do not let the question contain the explicit answer to the question. The question should not contain any direct references to the answer, and the answer should not be explicitly stated in the question. The question should be phrased in a way that requires the model to use reasoning to provide the correct answer.
- If the question has two parts, ensure that the second part does not simply reword the first part. The two parts should be distinctly worded and written.
- Assume that each question must stand on its own without previous questions. If you choose to generate two questions about the same topic, make sure that each question is self-contained and does not rely on the other question for context. Example: if the context is about a report, reference the report by name in any questions you generate.
- <IMPORTANT>DO NOT USE THE WORDS/PHRASES "THE CONTEXT", "THE DOCUMENT", "THE INFORMATION", "THE PROVIDED..." etc IN ANY QUESTION. THE TEST SET WILL NOT INCLUDE THE CONTEXT SO MODELS WILL NOT KNOW WHAT THAT MEANS AND THAT WOULD BE INEFFECTIVE.</IMPORTANT>
"""

\end{lstlisting}

\subsection{Correct and Hallucinated Answer Generation}
We used batch prompting to generate correct answers for each batch of generated questions.
\subsubsection{Correct Answer Generation Prompt}
\begin{lstlisting}[language=json,numbers=none]

"""Your role is to generate answers for a test set of evaluation questions generated from a knowledge base. You will be given a list of questions and a subsection of the knowledge base in the form of relevant context paragraphs to each question. The context paragraphs should include the information necessary to answer each question. Do not make up information that is not present in the context.


Your answer must be found in the provided context.  
Please respect the following rules to generate the question:
- The answer to the question should be found inside the provided context
- The question should be self-contained, hence the answer must be found in the context provided. Do not make up information that is not present in the context.
- The answer must be in this language: {language}

the "question_type" property will ALWAYS be one of the following: ['simple', 'reasoning', 'multi_context', 'situational', 'distracting', 'double', 'conditionals']. do not change this property during the answer generation, simply copy it from the input.

The user will provide the questions and the context, consisting of multiple paragraphs delimited by dashes "------".
You will append the precise answer to each question based exclusively on the provided context. Ensure that your answers are complete sentences that directly answer the question. Do not provide additional information that is not directly related to your answer to the question.
\n\n
You will be given an input of questions that have the following format: 
[{{"question": "xx1", "question_type":"<question type of xx1>"}}, {{"question": "xx2", "question_type":"<question type of xx2>"}}, ...]
as well as some context paragraphs.
You will output a list of {num_questions} JSON objects with keys 'question' mapping to the original question and 'answer' mapping to the answer you generate for the respective question, without any other wrapping text or markdown. Ensure that you will only return valid JSON, for example:
[{{"question": "xx1", "answer": "<answer to xx1>", "question_type":"<question type of xx1>"}}, {{"question": "xx2", "answer": "<answer to xx2>", "question_type":"<question type of xx2>"}}, ...].\n
"""

\end{lstlisting}

\subsubsection{Hallucinated Answer Generation Prompt}
\begin{lstlisting}[language=json,numbers=none]

"""Your role is to generate hallucinated answers for a given question from a knowledge base. You will be given a question and an answer, and your task is to create new answers that are plausible but do not necessarily align with the provided context.

The user will provide the question and the original answer.
You will return a list of {num_hallucinated_answers} JSON objects with keys 'question' and 'hallucinated_answer', without any other wrapping text or markdown. Ensure that you will only return valid JSON for example:
[{{"question": "xx", "hallucinated_answer": "yy", "question_type":"provided type of question xx"}}, {{"question": "xx1", "hallucinated_answer": "zz", "question_type":"provided type of question xx1"}}, ...].\n"""


\end{lstlisting}

\subsection{In-Context Learning}
\subsubsection{Verification Question Template Prompt}
We used a chain-of-verification prompting based on the prompts in \citep{ritunchainofverification2024}:
\begin{lstlisting}[language=json,numbers=none]

"""Your task is to create a verification question based on the below question provided.
Example Question: Who are some movie actors who were born in Boston?
Example Verification Question: Was [movie actor] born in [Boston]
Explanation: In the above example the verification question focused only on the ANSWER_ENTITY (name of the movie actor) and QUESTION_ENTITY (birth place).
Similarly you need to focus on the ANSWER_ENTITY and QUESTION_ENTITY from the actual question and generate verification question.

Actual Question: {original_question}

Final Verification Question:"""


\end{lstlisting}

\subsubsection{Verification Question Generation Prompt}
\begin{lstlisting}[language=json,numbers=none]

"""Your task is to create a series of verification questions based on the below question, the verification question template and baseline response.
Example Question: Who are some movie actors who were born in Boston?
Example Verification Question Template: Was [movie actor] born in Boston?
Example Baseline Response: Some movie actors born in Boston include: Matt Damon, Chris Evans.
Example Verification Question: 1. Was Matt Damon born in Boston?
2. Was Chris Evans born in Boston?

Explanation: In the above example the verification questions focused only on the ANSWER_ENTITY (name of the movie actor) and QUESTION_ENTITY (birth place) based on the template and substitutes entity values from the baseline response.
Similarly you need to focus on the ANSWER_ENTITY and QUESTION_ENTITY from the actual question and substitute the entity values from the baseline response to generate verification questions.

Actual Question: {original_question}
Baseline Response: {baseline_response}
Verification Question Template: {verification_question_template}

Final Verification Questions, separated by \\n:"""


\end{lstlisting}

\subsection{Filtering Suboptimal Questions}
\subsubsection{Filtering Criteria}
\label{question_filtering}
Part of the generation process involves filtering out questions that do not meet the criteria for an effective evaluation question. To optimize compute costs and minimize tokens used, we used a Regular Expression (RegEx) to match questions that contained certain sets of keywords that could imply ambiguity, or prevent the question from standing on its own without any additional context.

Our criteria for pattern matching was based on the following:
\begin{itemize}
    \item Direct references to an ambiguous "context", "document", "report", or "article".
    \item References to any day of the week, as some generated questions based on time-specific contexts (for example on news articles) do not specify the exact date of the event in question.
    \item Questions containing words such as "provided", "tasks", and "analyzed", that may indirectly refer to an ambiguous context.

\end{itemize}

We used the following RegEx pattern:

\begin{lstlisting}[language=json]
      r"(?i).*?(context|document|report|this|article|mentioned|in\s+the\s+context|in\s+the\s+document \\ |provided|above|average|f\s+value|compare|implications|tasks|was|analyzed|monday\\|tuesday |wednesday|thursday|friday|saturday|sunday|day|the).*?"
\end{lstlisting}

After selecting only the questions that matched the pattern, we batch prompted the high-performance LLM---in practice, \texttt{gpt-4o-mini}---to evaluate whether the questions selected were reasonable to include in an evaluation testset without additional context. We used the following few-shot learning \citep{brown2020languagemodelsfewshotlearners} prompt as shown in the next section. We stored the filtering model's \texttt{"reason"}s in a separate text file for manual review as needed.

\subsubsection{Question Filtering Prompt}
\begin{lstlisting}[language=json,numbers=none]
"""Your role is to filter questions generated by the question generation system. Your only criteria is to ensure that the questions are self contained, or in other words do not reference any unnamed or anonymous context that you do not see in the question itself. You will show your work by adding a 'valid' key to each question object in the array. If the question is missing a necessary piece of information or proper noun in order to be self contained, mark the valid key 'false'. If the question stands on its own, mark the valid key 'true'.

You will be given a list of questions in JSON format, and your task is to decide whether each question meets the criteria for being self contained. Include your reasoning for each decision in an additional 'reason' key added to each question object. We are not evaluating the model's ability to answer the question, only the question itself. Some models may be able to answer questions that reference historical events and academic studies without the need for expanded context - these are valid questions. However, questions that reference unnamed or anonymous context are not valid.


<EXAMPLES>
Example of a valid question:
'What is the capital of France?'
This is valid because the question can be answered based on general knowledge.

Example of a valid question:
'In the study conducted by Majolo et al., what was the initial pairing success rate of common marmoset females, and how does this rate compare to typical pairing success in other primate species?'
This is valid because the question is specific and unambiguous, with a clear and singular answer that can be found in the study, which is referenced by the authors' names and can be narrowed down to a specific study.

Example of a valid question:
'What is the relationship between message loss rates and estimation errors in the MDFU - LP technique, particularly in the context of system accuracy?'
This is a valid question because the context it references is defined - that being system accuracy. The question is specific and unambiguous, with a clear answer that can be reasoned out assuming the model has an understanding of the subject (MDFU-LP).

Example of a valid question:
'What is the significance of AIC in the context of statistical mathematics?'
This is a valid question because the context it references is defined - that being statistical mathematics. The question is specific and unambiguous, with a clear answer that can be reasoned out assuming that the model has an understanding of the subject (AIC).

Example of an invalid question:
'What is the author's favorite city?'
This is an invalid question because it is not clear who the author is, and the question cannot be answered without additional context that specifies who the author is.

Example of an invalid question:
'What type of behavior did younger monkeys show a higher attraction to in the study, and how does this attraction differ from the behaviors exhibited by older monkeys?'
This is an invalid question because it asks about a study without specifying which study, and the question cannot be answered without additional context that specifies the study being referenced.

Example of an invalid question:
"How many senior Russian political figures were included in the Treasury Department's list released on Monday?"
This is an invalid question because it is unclear which Monday is being referenced as the release date of the Treasury Department's list. Since the date is ambiguous, there could be multiple treasury department lists released on different Mondays, and the question cannot be answered without specifying the release date of the list.
</EXAMPLES>

For all valid questions in the JSON array, add a 'valid' key with a value of 'true' to the valid question object. 

Example input:
<EXAMPLE JSON>
[
    {{
        "question": "What is the capital of France?"
        "id": "1"
        ...(other keys - DO NOT TOUCH)
    }}, {{
        "question": "What is the author's favorite city?"
        "id": "2"
        ...(other keys - DO NOT TOUCH)
    }}
]
</EXAMPLE JSON>


Example output:
<EXAMPLE JSON>
[
    {{
        "question": "What is the capital of France?"
        "id": "1"
        "valid": "true"
        "reason": "This question is valid because it can be answered based on general knowledge."
        ...(other keys - DO NOT TOUCH)
    }},
    {{
        "question": "What is the author's favorite city?"
        "id": "2"
        "valid": "false"
        "reason": "This question is invalid because it is not clear who the author is, and the question cannot be answered without additional context that specifies who the author is."
        ...(other keys - DO NOT TOUCH)
    }}
]
</EXAMPLE JSON>
"""
\end{lstlisting}

\subsection{LoRA Parameters for Fine Tuning}
\label{lora_parameters}
Table \ref{tab:config1} shows the LoRA training configuration for fine-tuning in hallucination mitigation. 

\begin{table}[H]
\resizebox{\textwidth}{!}{%
\centering
\begin{tabular}{lll}
\hline
\textbf{Parameter Type} & \textbf{Description} &
\textbf{Config}\\
\hline
\texttt{r} & LoRA rank parameter & 8\\
\texttt{lora\_alpha} & LoRA alpha parameter & 16 \\
\texttt{bias} & Bias terms for LoRA & "None" \\
\texttt{lora\_dropout} & Dropout rate for LoRA & 0.05\\
\texttt{task\_type} & Task type, causal language modeling & "CAUSAL\_LM"\\
\texttt{warmup\_steps} & Warmup steps & 0\\
\texttt{per\_device\_train\_batch\_size} & Training batch size per device & 2\\
\texttt{gradient\_accumulation\_steps} & Number of steps for gradient accumulation & 2\\
\texttt{num\_train\_epochs} & Number of training epochs & 3\\
\texttt{learning\_rate} & Learning rate & 2e-5\\
\texttt{optim} & Optimizer type & "paged\_adamw\_8bit"\\
\texttt{logging\_steps} & Number of steps for logging & 25\\
\texttt{save\_steps} & Number of steps between saving & 1\\
\texttt{eval\_steps} & Number of steps between evaluations & 25\\
\texttt{do\_eval} & Enable evaluation & True \\
\texttt{gradient\_checkpointing} & Enable gradient checkpointing & True\\
\hline
\end{tabular}
}
\caption{Quantization and LoRA Configuration}
\label{tab:config1}
\end{table}

\subsection{Text Generation Scoring Metrics}
 \label{additional formulas}
We used the following formulae for calculating text generation evaluation metrics. They are based on the metrics detailed in the RAGAS \citep{es2023ragasautomatedevaluationretrieval} framework.

\begin{itemize}
    \item Answer Faithfulness
    \item Answer Similarity
    \item Answer Correctness
    \item Factual Correctness
\end{itemize}

\subsubsection{Answer Faithfulness Calculation}
The formula for calculating Answer Faithfulness ($A_F$) is:
\begin{equation}
    A_F = \frac{|V|}{|S|}
\end{equation}
where 
\begin{itemize}
    \item $A_F$ means the faithfulness score.
    \item |V| means the number of statements that are supported by LLM.
    \item |S| means the number of total statements.
\end{itemize}

\subsubsection{Answer Similarity Calculation}
Answer Similarity ($A_S$) is calculated as the  cosine similarity between two vectors:
\begin{equation}
    \texttt{Answer Similarity} = \frac{\mathbf{V_C} \cdot \mathbf{V_G}}{||\mathbf{V_C}||||\mathbf{V_G}||}
\end{equation}
where
\begin{itemize}
    \item $V_C$ means the embedding vector of the correct answer using the specified embedding model.
    \item $V_G$ means the embedding vector of the generated answer using hte specified embedding model.
\end{itemize}

\subsubsection{Answer Correctness Calculation}
Answer Correctness ($A_C$) is computed by the weighted average of factual similarity and semantic similarity ($A_S$):
\begin{equation}
    A_C = \frac{(w_1 \cdot \texttt{Factual Correctness}) + (w_2 \cdot A_S)}{w_1 + w_2}
\end{equation}

The weight parameter $w_1$ are set to 0.75 as default and $w_2 = 1 - w_1 = 0.25$. The factual correctness computes the factual overlap between the generated answer and the ground truth answer, and its calculated as:

\begin{equation}
    \texttt{Factual Correctness} = \frac{|\text{TP}|}{|\text{TP}| + 0.5 \times (|\text{FP}| + |\text{FN}|)}
\end{equation}
where
\begin{itemize}
    \item TP (True Positive): Statements that are both shown in the ground truth and the generated answer.
    \item FP (False Positive): Statements that are shown in the generated answer but not in the ground truth.
    \item FN (False Negative): Statements that are shown in the ground truth but not in the generated answer.
\end{itemize}

\subsection{Text Node Sampling Analysis} \label{sampling}
\begin{figure}[H]
    \centering
    \includegraphics[width=1\linewidth]{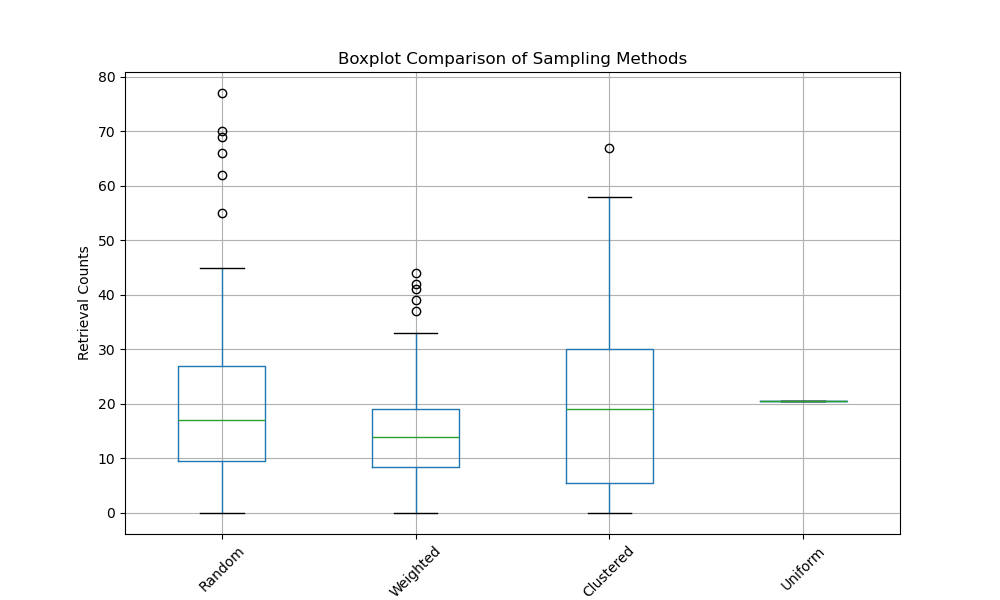}
    \caption{Box Plot showing Uniform Effectiveness of Different Sampling Methods}
    \label{fig:sampling1}
\end{figure}

\begin{figure}[H]
    \centering    \includegraphics[width=1\linewidth]{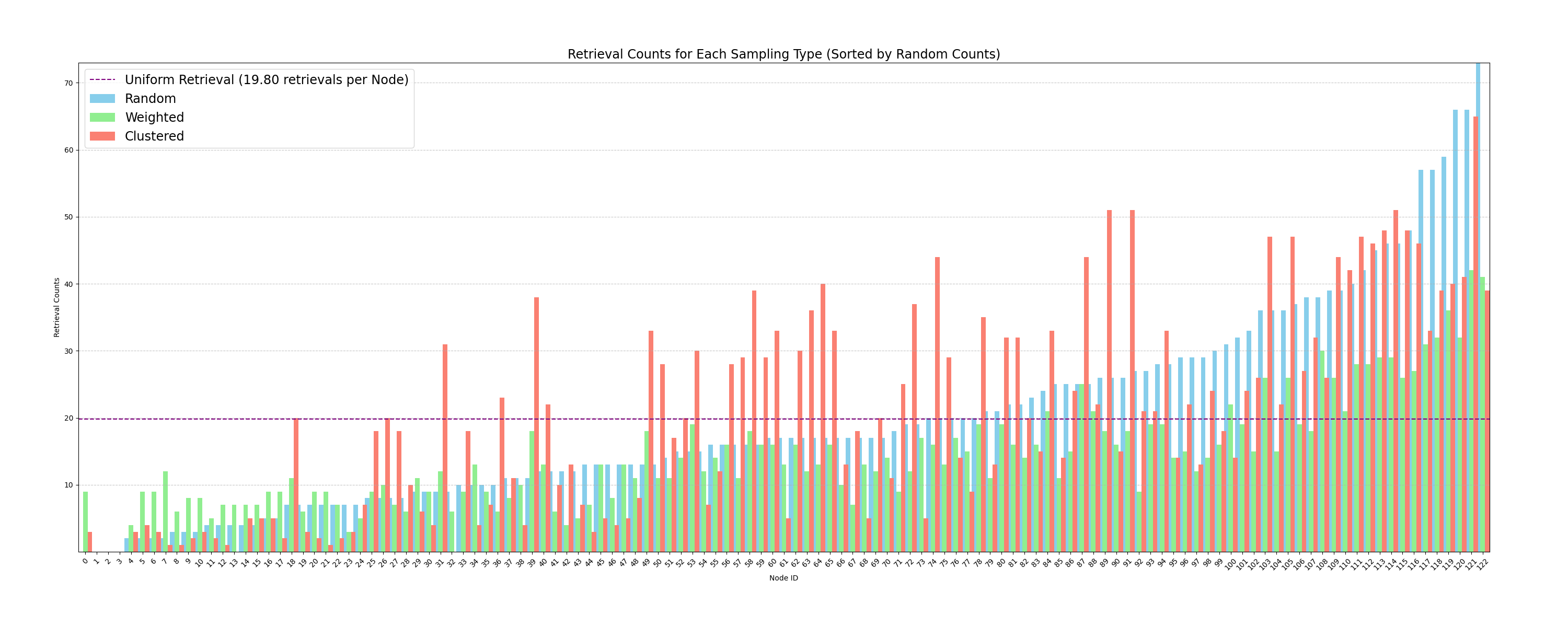}
    \caption{Retrieval Counts by Node ID}
    \label{fig:sampling2}
\end{figure}

Both figure \ref{fig:sampling1} and \ref{fig:sampling2} shows the text node retrieval distribution comparison of different sampling methods. The results show that the weighted sampling method we designed has a significant improvement over random sampling, and node retrieval distribution is closer to a uniform distribution, indicating that our dataset covers the original corpus more evenly.


\subsection{Model Comparison for Different Question Types}

\begin{table}[h]
  \caption{Comparison of models on Hallucination Generation and Detection for "Simple" Questions}
  \label{table1}
  \centering
  \begin{tabular}{lcccccccc}
    \toprule
    Model & \multicolumn{4}{c}{Text Generation} & \multicolumn{4}{c}{Hallucination Identification}\\
    \cmidrule(r){2-5} \cmidrule(r){6-9}
   & $A_F$     & $A_R$     & $A_C$ & $A_S$ & Acc. & Prec. & Rec. & F1
   \\
    \midrule
    GPT-4o (13-05-2024) & & & & & & & &
    \\
    \midrule
    Original   & 0.484 & 0.778 & 0.670   & 0.767  & 0.562& 0.529 & 0.586& 0.556
   \\
    ICL     & 0.512 & 0.773 & 0.668 & 0.768 & 0.600 & 0.605 & 0.640 & 0.622   \\
    RAG  & 0.494 & 0.787 & 0.674 & 0.768 & 0.558 & 0.505 & 0.269 & 0.351  \\
    \midrule
    GPT-4o-mini & & & & & & & &
    \\
    \midrule
    Original   & 0.426 & 0.826 & 0.670 & 0.801 & 0.565 & 0.720 & 0.171 & 0.276  \\
    ICL  & 0.422 & 0.823 & 0.667 & 0.801 & 0.514 & 0.691 & 0.164 & 0.265  \\
    RAG  & 0.400 & 0.813 & 0.652 & 0.803 & 0.643 & 0.689 & 0.590 & 0.635  \\
    \midrule
    Mistral-Nemo & & & & & & & &
    \\
    \midrule
    Original & 0.275 & 0.439 & 0.130 & 0.075 & 0.447 & 0.409 & 0.286 & 0.337  \\
    ICL & 0.298 & 0.424 & 0.124 & 0.075 & 0.446 & 0.378 & 0.263 & 0.310   \\
    RAG & 0.306 & 0.413 & 0.130 & 0.075 & 0.516 & 0.589 & 0.293 & 0.391   \\
    \midrule
    Llama3.1-8b & & & & & & & &
    \\
    \midrule
    Original& 0.375 & 0.496 & 0.475 & 0.707 & 0.488 & 0.482 & 0.682 & 0.565  \\
    ICL & 0.378 & 0.498 & 0.474 & 0.701 & 0.536 & 0.562 & 0.736 & 0.638 \\
    RAG & 0.380 & 0.493 & 0.481 & 0.696 & 0.526 & 0.535 & 0.873 & 0.663 \\
    PEFT & 0.227 & 0.790 & 0.531 & 0.705 & 0.483 & 0.482 & 1.000 & 0.656 \\
    \bottomrule
  \end{tabular}
\end{table}

\begin{table}[h]
  \caption{Comparison of models on Hallucination Generation and Detection for "Reasoning" Questions}
  \label{table2}
  \centering
  \begin{tabular}{lcccccccc}
    \toprule
    Model & \multicolumn{4}{c}{Text Generation} & \multicolumn{4}{c}{Hallucination Identification}\\
    \cmidrule(r){2-5} \cmidrule(r){6-9}
   & $A_F$     & $A_R$     & $A_C$ & $A_S$ & Acc. & Prec. & Rec. & F1\\
    \midrule
    GPT-4o & & & & & & & &
    \\
    \midrule
    Original   & 0.720 & 0.717 & 0.755 & 0.783 & 0.660 & 0.707 & 0.451 & 0.550
   \\
    ICL     & 0.687 & 0.701 & 0.774 & 0.782 & 0.629 & 0.733 & 0.436 & 0.547  \\
    RAG  & 0.696 & 0.706 & 0.767 & 0.782 & 0.685 & 0.870 & 0.461 & 0.603  \\
    \midrule
    GPT-4o-mini & & & & & & & &
    \\
    \midrule
    Original   & 0.575 & 0.721 & 0.782 & 0.794 & 0.624 & 0.933 & 0.161 & 0.275  \\
    ICL  & 0.574 & 0.725 & 0.768 & 0.794 & 0.589 & 1.000 & 0.120 & 0.214  \\
    RAG  & 0.546 & 0.701 & 0.766 & 0.799 & 0.609 & 0.727 & 0.392 & 0.510  \\
    \midrule
    Mistral-Nemo & & & & & & & &
    \\
    \midrule
    Original & 0.497 & 0.444 & 0.154 & 0.073 & 0.437 & 0.393 & 0.222 & 0.284 \\
    ICL & 0.434 & 0.443 & 0.151 & 0.073 & 0.472 & 0.321 & 0.214 & 0.257   \\
    RAG & 0.459 & 0.444 & 0.154 & 0.073 & 0.481 & 0.481 & 0.241 & 0.321   \\
    \midrule
    Llama3.1-8b & & & & & & & &
    \\
    \midrule
    Original& 0.490 & 0.609 & 0.654 & 0.748 & 0.540 & 0.567 & 0.628 & 0.596  \\
    ICL & 0.548 & 0.617 & 0.673 & 0.753 & 0.600 & 0.643 & 0.656 & 0.649 \\
    RAG & 0.531 & 0.621 & 0.670 & 0.748 & 0.457 & 0.442 & 0.913 & 0.596 \\
    PEFT & 0.312 & 0.652 & 0.643 & 0.765 & 0.457 & 0.457 & 1.000 & 0.627 \\
    \bottomrule
  \end{tabular}
\end{table}

\begin{table}[h]
  \caption{Comparison of models on Hallucination Generation and Detection for "Multi Context" Questions}
  \label{table3}
  \centering
  \begin{tabular}{lcccccccc}
    \toprule
    Model & \multicolumn{4}{c}{Text Generation} & \multicolumn{4}{c}{Hallucination Identification}\\
    \cmidrule(r){2-5} \cmidrule(r){6-9}
   & $A_F$     & $A_R$     & $A_C$ & $A_S$ & Acc. & Prec. & Rec. & F1\\
    \midrule
    GPT-4o & & & & & & & &
    \\
    \midrule
    Original   & 0.654 & 0.762 & 0.752 & 0.807 & 0.649 & 0.635 & 0.784 & 0.702
   \\
    ICL     & 0.656 & 0.756 & 0.736 & 0.807 & 0.644 & 0.639 & 0.757 & 0.693  \\
    RAG  & 0.700 & 0.743 & 0.746 & 0.805 & 0.711 & 0.803 & 0.552 & 0.654 \\
    \midrule
    GPT-4o-mini & & & & & & & &
    \\
    \midrule
    Original   & 0.534 & 0.762 & 0.727 & 0.817 & 0.598 & 0.667 & 0.286 & 0.400  \\
    ICL  & 0.513 & 0.767 & 0.733 & 0.814 & 0.660 & 0.647 & 0.289 & 0.400  \\
    RAG  & 0.454 & 0.748 & 0.715 & 0.815 & 0.660 & 0.754 & 0.515 & 0.612  \\
    \midrule
    Mistral-Nemo & & & & & & & &
    \\
    \midrule
    Original & 0.293 & 0.371 & 0.230 & 0.061 & 0.418 & 0.393 & 0.240 & 0.298 \\
    ICL & 0.291 & 0.368 & 0.234 & 0.061 & 0.412 & 0.400 & 0.280 & 0.329  \\
    RAG & 0.304 & 0.360 & 0.230 & 0.061 & 0.503 & 0.500 & 0.315 & 0.386   \\
    \midrule
    Llama3.1-8b & & & & & & & &
    \\
    \midrule
    Original& 0.460 & 0.433 & 0.516 & 0.702 & 0.539 & 0.504 & 0.747 & 0.602  \\
    ICL & 0.470 & 0.425 & 0.528 & 0.673 & 0.562 & 0.577 & 0.674 & 0.621 \\
    RAG & 0.447 & 0.427 & 0.501 & 0.682 & 0.483 & 0.484 & 0.929 & 0.637 \\
    PEFT & 0.241 & 0.744 & 0.640 & 0.751 & 0.457 & 0.457 & 1.000 & 0.627 \\
    \bottomrule
  \end{tabular}
\end{table}

\begin{table}[h]
  \caption{Comparison of models on Hallucination Generation and Detection for "Distracting" Questions}
  \label{table4}
  \centering
  \begin{tabular}{lcccccccc}
    \toprule
    Model & \multicolumn{4}{c}{Text Generation} & \multicolumn{4}{c}{Hallucination Identification}\\
    \cmidrule(r){2-5} \cmidrule(r){6-9}
   & $A_F$  & $A_R$     & $A_C$ & $A_S$ & Acc. & Prec. & Rec. & F1\\
    \midrule
    GPT-4o & & & & & & & &
    \\
    \midrule
    Original   & 0.435 & 0.755 & 0.693 & 0.765 & 0.635 & 0.597 & 0.769 & 0.672
   \\
    ICL     & 0.425 & 0.757 & 0.695 & 0.765 & 0.598 & 0.565 & 0.720 & 0.633 \\
    RAG  & 0.437 & 0.760 & 0.692 & 0.765 & 0.634 & 0.777 & 0.497 & 0.606 \\
    \midrule
    GPT-4o-mini & & & & & & & &
    \\
    \midrule
    Original   & 0.380 & 0.779 & 0.693 & 0.781 & 0.586 & 0.742 & 0.265 & 0.390 \\
    ICL  & 0.376 & 0.786 & 0.682 & 0.781 & 0.604 & 0.731 & 0.264 & 0.387 \\
    RAG  & 0.347 & 0.758 & 0.673 & 0.776 & 0.711 & 0.740 & 0.752 & 0.746  \\
    \midrule
    Mistral-Nemo & & & & & & & &
    \\
    \midrule
    Original & 0.266 & 0.396 & 0.154 & 0.082 & 0.403 & 0.363 & 0.313 & 0.336 \\
    ICL & 0.276 & 0.389 & 0.156 & 0.082 & 0.455 & 0.368 & 0.320 & 0.342 \\
    RAG & 0.274 & 0.386 & 0.157 & 0.083 & 0.489 & 0.521 & 0.250 & 0.338  \\
    \midrule
    Llama3.1-8b & & & & & & & &
    \\
    \midrule
    Original& 0.386 & 0.379 & 0.435 & 0.672 & 0.501 & 0.497 & 0.736 & 0.593 \\
    ICL & 0.354 & 0.392 & 0.441 & 0.672 & 0.571 & 0.552 & 0.796 & 0.652 \\
    RAG & 0.380 & 0.390 & 0.449 & 0.665 & 0.522 & 0.496 & 0.805 & 0.614 \\
    PEFT & 0.223 & 0.761 & 0.578 & 0.711 & 0.540 & 0.540 & 1.000 & 0.701 \\
    \bottomrule
  \end{tabular}
\end{table}

\begin{table}[h]
  \caption{Comparison of models on Hallucination Generation and Detection for "Situational" Questions}
  \label{table5}
  \centering
  \begin{tabular}{lcccccccc}
    \toprule
    Model & \multicolumn{4}{c}{Text Generation} & \multicolumn{4}{c}{Hallucination Identification}\\
    \cmidrule(r){2-5} \cmidrule(r){6-9}
   & $A_F$ & $A_R$ & $A_C$ & $A_S$ & Acc. & Prec. & Rec. & F1\\
    \midrule
    GPT-4o & & & & & & & &
    \\
    \midrule
    Original   & 0.522 & 0.785 & 0.672 & 0.784 & 0.590 & 0.583 & 0.773 & 0.665
   \\
    ICL     & 0.511 & 0.789 & 0.678 & 0.785 & 0.576 & 0.553 & 0.800 & 0.654 \\
    RAG  & 0.517 & 0.794 & 0.680 & 0.786 & 0.592 & 0.632 & 0.426 & 0.509\\
    \midrule
    GPT-4o-mini & & & & & & & &
    \\
    \midrule
    Original   & 0.396 & 0.812 & 0.666 & 0.798 & 0.580 & 0.728 & 0.258 & 0.381  \\
    ICL  & 0.405 & 0.817 & 0.669 & 0.797 & 0.592 & 0.704 & 0.233 & 0.350 \\
    RAG  & 0.378 & 0.800 & 0.662 & 0.797 & 0.728 & 0.736 & 0.755 & 0.745  \\
    \midrule
    Mistral-Nemo & & & & & & & &
    \\
    \midrule
    Original & 0.311 & 0.437 & 0.126 & 0.080 & 0.461 & 0.426 & 0.348 & 0.383 \\
    ICL & 0.307 & 0.435 & 0.127 & 0.080 & 0.459 & 0.415 & 0.369 & 0.390  \\
    RAG & 0.293 & 0.436 & 0.126 & 0.080 & 0.524 & 0.479 & 0.272 & 0.347  \\
    \midrule
    Llama3.1-8b & & & & & & & &
    \\
    \midrule
    Original& 0.386 & 0.453 & 0.475 & 0.707 & 0.576 & 0.577 & 0.846 & 0.686  \\
    ICL & 0.404 & 0.468 & 0.480 & 0.694 & 0.499 & 0.513 & 0.780 & 0.619 \\
    RAG & 0.413 & 0.476 & 0.504 & 0.700 & 0.530 & 0.530 & 0.937 & 0.677 \\
    PEFT & 0.214 & 0.771 & 0.530 & 0.691 & 0.513 & 0.513 & 1.000 & 0.678 \\
    \bottomrule
  \end{tabular}
\end{table}

\begin{table}[h]
  \caption{Comparison of models on Hallucination Generation and Detection for "Double" Questions}
  \label{table6}
  \centering
  \begin{tabular}{lcccccccc}
    \toprule
    Model & \multicolumn{4}{c}{Text Generation} & \multicolumn{4}{c}{Hallucination Identification}\\
    \cmidrule(r){2-5} \cmidrule(r){6-9}
   & $A_F$ & $A_R$ & $A_C$ & $A_S$ & Acc. & Prec. & Rec. & F1\\
    \midrule
    GPT-4o & & & & & & & &
    \\
    \midrule
    Original   & 0.546 & 0.725 & 0.709 & 0.856 & 0.588 & 0.539 & 0.890 & 0.672
   \\
    ICL     & 0.549 & 0.713 & 0.706 & 0.857 & 0.597 & 0.575 & 0.901 & 0.702 \\
    RAG  & 0.553 & 0.720 & 0.702 & 0.854 & 0.682 & 0.648 & 0.707 & 0.676 \\
    \midrule
    GPT-4o-mini & & & & & & & &
    \\
    \midrule
    Original   & 0.469 & 0.721 & 0.637 & 0.863 & 0.621 & 0.674 & 0.323 & 0.437 \\
    ICL  & 0.422 & 0.729 & 0.646 & 0.862 & 0.621 & 0.776 & 0.355 & 0.487 \\
    RAG  & 0.396 & 0.729 & 0.649 & 0.857 & 0.725 & 0.741 & 0.754 & 0.748  \\
    \midrule
    Mistral-Nemo & & & & & & & &
    \\
    \midrule
    Original & 0.358 & 0.377 & 0.245 & 0.073 & 0.450 & 0.476 & 0.348 & 0.402 \\
    ICL & 0.349 & 0.372 & 0.245 & 0.073 & 0.483 & 0.466 & 0.327 & 0.384  \\
    RAG & 0.334 & 0.357 & 0.242 & 0.073 & 0.521 & 0.474 & 0.321 & 0.383  \\
    \midrule
    Llama3.1-8b & & & & & & & &
    \\
    \midrule
    Original& 0.319 & 0.418 & 0.442 & 0.714 & 0.535 & 0.524 & 0.667 & 0.587  \\
    ICL & 0.331 & 0.396 & 0.430 & 0.713 & 0.592 & 0.563 & 0.784 & 0.655 \\
    RAG & 0.320 & 0.388 & 0.432 & 0.710 & 0.531 & 0.510 & 0.891 & 0.649 \\
    PEFT & 0.227 & 0.690 & 0.549 & 0.802 & 0.488 & 0.488 & 1.000 & 0.656 \\
    \bottomrule
  \end{tabular}
\end{table}

\end{document}